%% file: main.tex
\documentclass{article} 
\usepackage{colm2024_conference}

\usepackage{microtype}
\usepackage{hyperref}
\usepackage{url}
\usepackage{booktabs}
\definecolor{darkblue}{rgb}{0, 0, 0.5}
\hypersetup{colorlinks=true, citecolor=darkblue, linkcolor=darkblue, urlcolor=darkblue}

\usepackage{enumitem}
\usepackage{graphicx}
\usepackage{url}
\usepackage{bigfoot}
\usepackage{booktabs}
\usepackage{multirow}
\usepackage[most]{tcolorbox}
\usepackage{graphicx}
\usepackage{subcaption}
\usepackage{etoolbox}
\usepackage{booktabs}
\usepackage{bbding} 
\let\classAND\AND
\let\AND\relax
\usepackage{algorithmic}

\let\AND\classAND
\AtBeginEnvironment{algorithmic}{\let\AND\algoAND}

\usepackage{algorithm}
\usepackage{algorithmic}

\usepackage{soul}
\usepackage{url}
\usepackage{booktabs}
\usepackage{arydshln}
\usepackage{nicefrac}
\usepackage{multirow}
\usepackage{tabularx}
\usepackage{amsmath}
\usepackage{amssymb}
\usepackage{amsfonts}
\usepackage{mathtools}
\usepackage{fontawesome}

\usepackage{xcolor,colortbl}
\usepackage[export]{adjustbox}

\usepackage{wrapfig}

\tcbuselibrary{listings,breakable}
\tcbset{listing engine=listings,colframe=black,colback=white,size=small}

\usepackage{upquote}
\definecolor{dkgreen}{rgb}{0,0.6,0}
\definecolor{gray}{rgb}{0.5,0.5,0.5}
\definecolor{mauve}{rgb}{0.58,0,0.82}
\usepackage{xspace}

\usepackage{setspace}

\tcbset{
  promptstyle/.style={
    colback=gray!15,   
    colframe=gray!60,  
    boxrule=0.4pt,     
    arc=2pt,           
    left=4pt,right=4pt,top=4pt,bottom=4pt,  
    fonttitle=\bfseries,
  }
}
\newtcolorbox{promptblock}[1][]{promptstyle,#1}

\usepackage{fvextra}
\RecustomVerbatimEnvironment{verbatim}{Verbatim}{breaklines=true, breakanywhere=true}

\newcommand{\huggingface}{\raisebox{-1.5pt}{\includegraphics[height=1.05em]{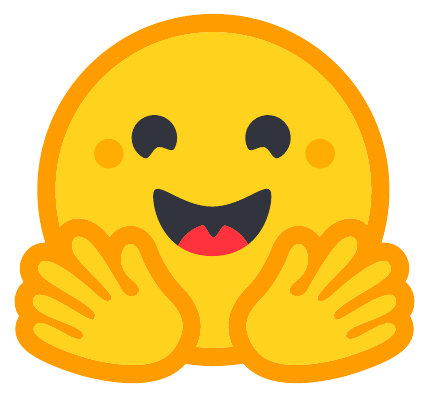}}\hspace{3.2pt}\xspace}
\newcommand{\modelscope}{\raisebox{-1.5pt}{\includegraphics[height=1.05em]{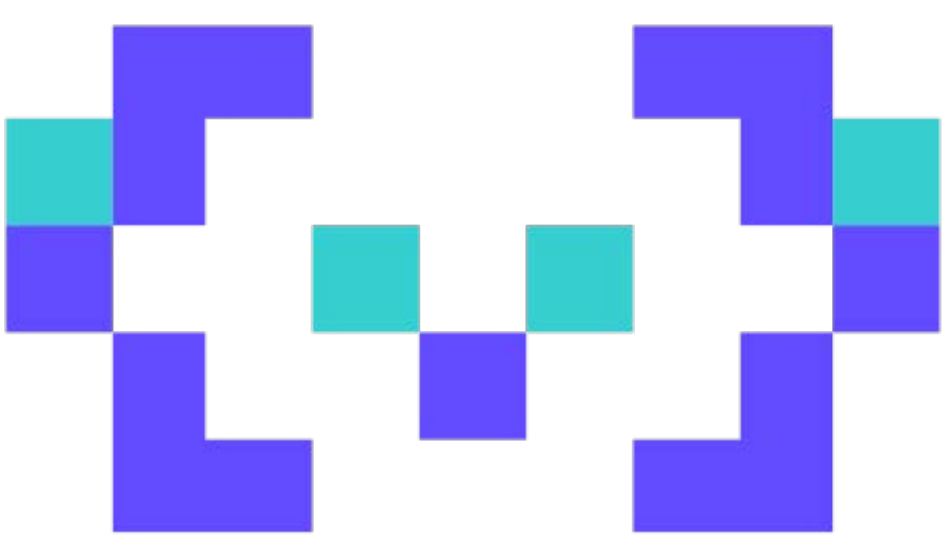}}\xspace}
\newcommand{\github}{\raisebox{-1.5pt}{\includegraphics[height=1.05em]{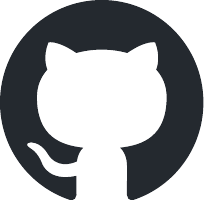}}\hspace{3.2pt}\xspace}

\newcommand\blfootnote[1]{
    \begingroup
    \renewcommand\thefootnote{}\footnote{#1}
    \addtocounter{footnote}{-1}
    \endgroup
}

\title{\LARGE Qwen3 Embedding: Advancing Text Embedding and Reranking Through Foundation Models}

\author{
\vspace{-4mm}
\\
\normalsize{}
Yanzhao Zhang*\hspace{3mm} 
Mingxin Li*\hspace{3mm} 
Dingkun Long*\hspace{3mm} 
Xin Zhang*\hspace{3mm} 
\\
\normalsize{}
Huan Lin\hspace{3mm} 
Baosong Yang\hspace{3mm}
Pengjun Xie\hspace{3mm}
An Yang\hspace{3mm} 
\\
\normalsize{}
Dayiheng Liu\hspace{3mm} 
Junyang Lin\hspace{3mm} 
Fei Huang\hspace{3mm}
Jingren Zhou\hspace{3mm}
\\
\vspace{20mm}
\textbf{Tongyi Lab\hspace{3mm}Alibaba Group}
\vspace{-15mm}
}

\colmfinalcopy
\begin{document}

\maketitle

\begin{center}
\vspace{-1cm}
\begin{tabular}{rl}
\huggingface & \url{https://huggingface.co/Qwen}\\
\modelscope & \url{https://modelscope.cn/organization/qwen} \\
\github & \url{https://github.com/QwenLM/Qwen3-Embedding}\\
\end{tabular}
\end{center}

\begin{abstract}
In this work, we introduce the Qwen3 Embedding series, a significant advancement over its predecessor, the GTE-Qwen series, in text embedding and reranking capabilities, built upon the Qwen3 foundation models. Leveraging the Qwen3 LLMs' robust capabilities in multilingual text understanding and generation, our innovative multi-stage training pipeline combines large-scale unsupervised pre-training with supervised fine-tuning on high-quality datasets. Effective model merging strategies further ensure the robustness and adaptability of the Qwen3 Embedding series. During the training process, the Qwen3 LLMs serve not only as backbone models but also play a crucial role in synthesizing high-quality, rich, and diverse training data across multiple domains and languages, thus enhancing the training pipeline. The Qwen3 Embedding series offers a spectrum of model sizes (0.6B, 4B, 8B) for both embedding and reranking tasks, addressing diverse deployment scenarios where users can optimize for either efficiency or effectiveness.
Empirical evaluations demonstrate that the Qwen3 Embedding series achieves state-of-the-art results across diverse benchmarks. Notably, it excels on the multilingual evaluation benchmark MTEB for text embedding, as well as in various retrieval tasks, including code retrieval, cross-lingual retrieval and multilingual retrieval. To facilitate reproducibility and promote community-driven research and development, the Qwen3 Embedding models are publicly available under the Apache 2.0 license.
\end{abstract}

\blfootnote{$^{*}$ Equal contribution}

\input{contents/intro}
\input{contents/model_arc}
\input{contents/train}
\input{contents/evaluation}

\section{Conclusion}
In this technical report, we present the Qwen3-Embedding series, a comprehensive suite of text embedding and reranking models based on the Qwen3 foundation models. These models are designed to excel in a wide range of text embedding and reranking tasks, including multilingual retrieval, code retrieval, and complex instruction following. The Qwen3-Embedding models are built upon a robust multi-stage training pipeline that combines large-scale weakly supervised pre-training on synthetic data with supervised fine-tuning and model merging on high-quality datasets. The Qwen3 LLMs play a crucial role in synthesizing diverse training data across multiple languages and tasks, thereby enhancing the models' capabilities. Our comprehensive evaluations demonstrate that the Qwen3-Embedding models achieve state-of-the-art performance across various benchmarks, including MTEB, CMTEB, MMTEB, and several retrieval benchmarks. We are pleased to open-source the Qwen3-Embedding and Qwen3-Reranker models (0.6B, 4B, and 8B), making them available for the community to use and build upon.

\bibliography{colm2024_conference}
\bibliographystyle{colm2024_conference}

\clearpage
\appendix
\section{Appendix}

\subsection{Synthetic Data} \label{synthetic_data}
We construct four types of synthetic data—retrieval, bitext mining, semantic textual similarity, and classification to enable the model to adapt to various similarity tasks during pre-training. To ensure both multilingual and cross-lingual diversity, the data is generated using Qwen3 32B. Below is an example of a synthetic retrieval text pair. The retrieval data is synthesized using a document-to-query approach. We collect a multilingual corpus from the pre-training corpus of the Qwen3 base model to serve as the document source. A two-stage generation pipeline is then applied, consisting of: (1) configuration and (2) query generation. In the configuration stage, we use large language models (LLMs) to determine the ``Question Type'', ``Difficulty'', and ``Character'' for the synthetic query. The candidate characters are retrieved from Persona Hub~\citep{ge2024scaling}, selecting the top five most relevant to the given document. This step aims to enhance the diversity of the generated queries. The template used is as follows:

\begin{promptblock}
    \begin{verbatim}
Given a **Passage** and **Character**, select the appropriate option from three fields: Character, Question_Type, Difficulty, and return the output in JSON format.
First, select the Character who are likely to be interested in the Passage from the candidates. Then select the Question_Type that the Character might ask about the Passage; Finally, choose the Difficulty of the possible question based on the Passage, the Character, and the Question_Type.
Character: Given by input **Character**

Question_Type:
- keywords: ...
- acquire_knowledge: ...
- summary: ...
- yes_or_no: ...
- background: ...

Difficulty:
- high_school: ...
- university: ...
- phd: ...

Here are some examples
<Example1> <Example2> <Example3>

Now, generate the **output** based on the **Passage** and **Character** from user, the **Passage** will be in {language} language and the **Character** will be in English.
Ensure to generate only the JSON output with content in English.

**Passage**:
{passage}
**Character**:
{character}
\end{verbatim}
\end{promptblock}

In the query generation stage, we use the configuration selected in the first stage to guide the generation of queries. Additionally, we explicitly specify the desired length and language of the generated query. The template used is as follows:

\begin{promptblock}
    \begin{verbatim}
Given a **Character**, **Passage**, and **Requirement**, generate a query from the **Character**'s perspective that satisfies the **Requirement** and can be used to retrieve the **Passage**. Please return the result in JSON format.

Here is an example:
<example>

Now, generate the **output** based on the **Character**, **Passage** and **Requirement** from user, the **Passage** will be in {corpus_language} language, the **Character** and **Requirement** will be in English.
Ensure to generate only the JSON output, with the key in English and the value in {queries_language} language.

**Character**
{character}
**Passage**
{passage}
**Requirment**
- Type: {type};
- Difficulty: {difficulty};
- Length: the length of the generated sentences should be {length} words;
- Languange: the language in which the results are generated should be {language} language;
    \end{verbatim}
\end{promptblock}

\input{tables/train_data}

\subsection{Detail Results}

\input{tables/mteb_en.tex}
\input{tables/mteb_zh.tex}
\input{tables/mteb_code.tex}

\end{document}

%% file: contents/intro.tex
\section{Introduction}
\label{sec:intro}
Text embedding and reranking are fundamental components in numerous natural language processing and information retrieval applications, including web search, question answering, recommendation systems, and beyond~\citep{karpukhin2020dense,huang2020embedding,zhao2023embedding,zhao2024dense}. High-quality embeddings enable models to capture semantic relationships between texts, while effective reranking mechanisms ensure that the most relevant results are prioritized. Recently, emerging application paradigms such as Retrieval-Augmented Generation (RAG) and agent systems, driven by the advancement of large language models (e.g., Qwen3 \citep{yang2025qwen3}, GPT-4o \citep{hurst2024gpt}), have introduced new requirements and challenges for text embedding and reranking, both in terms of model training paradigms and application scenarios. Despite significant advancements, training embedding and reranking models that perform well in scalability, contextual understanding, and alignment with specific downstream tasks remains challenging.

The emergence of large language models (LLMs) has significantly advanced the development of text embedding and reranking models. Prior to the introduction of LLMs, the predominant approach involved using encoder-only pretrained language models like BERT as the foundational model for training \citep{reimers-gurevych-2019-sentence}. The richer world knowledge, text understanding, and reasoning abilities inherent in LLMs have led to further enhancements in models trained on these architectures. Additionally, there has been considerable research facilitating the integration of LLMs into processes such as training data synthesis and quality data filtering \citep{wang-etal-2024-improving-text,lee2024nv,lee2025gemini}. The fundamental characteristics of LLMs have also inspired the introduction of new training paradigms. For instance, during the embedding model training process, incorporating differentiated tasks across aspects such as instruction type, domain, and language has yielded improved performance in downstream tasks \citep{su2023one}. Similarly, for reranking model training, advancements have been realized through both zero-shot methods based on user prompts and approaches combining supervised fine-tuning \citep{ma2023zero,pradeep2023rankvicuna,zhang2024two,zhuang2024setwise}.

In this work, we introduce the Qwen3 Embedding series models, which are constructed on top of the Qwen3 foundation models. The Qwen3 foundation has simultaneously released base and instruct model versions, and we exploit the robust multilingual text understanding and generation capabilities of these models to fully realize their potential in training embedding and reranking models. To train the embedding models, we implement a multi-stage training pipeline that involves large-scale unsupervised pre-training followed by supervised fine tuning on high-quality datasets. We also employ model merging with various model checkpoints to enhance robustness and generalization. The Qwen3 instruct model allows for efficient synthesis of a vast, high-quality, multilingual, and multi-task text relevance dataset. This synthetic data is utilized in the initial unsupervised training stage, while a subset of high-quality, small-scale data is selected for the second stage of supervised training. For the reranking models, we adopt a two-stage training scheme in a similar manner, consisting of high-quality supervised fine tuning and a model merging stage. Based on different sizes of the Qwen3 backbone models (including 0.6B, 4B, and 8B), we ultimately trained three text embedding models and three text reranking models. To facilitate their application in downstream tasks, the Qwen3 Embedding series supports several practical features, such as flexible dimension representation for embedding models and customizable instructions for both embedding and reranking models.

We evaluate the Qwen3 Embedding series across a comprehensive set of benchmarks spanning multiple tasks and domains. Experimental results demonstrate that our embedding and reranking models achieve state-of-the-art performance, performing competitively against leading proprietary models in several retrieval tasks. For example, the flagship model Qwen3-8B-Embedding attains a score of 70.58 on the MTEB Multilingual benchmark~\citep{enevoldsen2025mmteb} and 80.68 on the MTEB Code benchmark~\citep{enevoldsen2025mmteb}, surpassing the previous state-of-the-art proprietary embedding model, Gemini-Embedding~\citep{lee2025gemini}. Moreover, our reranking model delivers competitive results across a range of retrieval tasks. The Qwen3-Reranker-0.6B model exceeds previously top-performing models in numerous retrieval tasks, while the larger Qwen3-Reranker-8B model demonstrates even superior performance, improving ranking results by 3.0 points over the 0.6B model across multiple tasks. Furthermore, we include a constructive ablation study to elucidate the key factors contributing to the superior performance of the Qwen3 Embedding series, providing insights into its effectiveness.

In the following sections, we describe the design of the model architecture, detail the training procedures, present the experimental results for both the embedding and reranking models of the Qwen3 Embedding Series, and conclude this technical report by summarizing the key findings and outlining potential directions for future research.

%% file: contents/model_arc.tex
\begin{figure}
\centering
\includegraphics[width=0.9\linewidth]{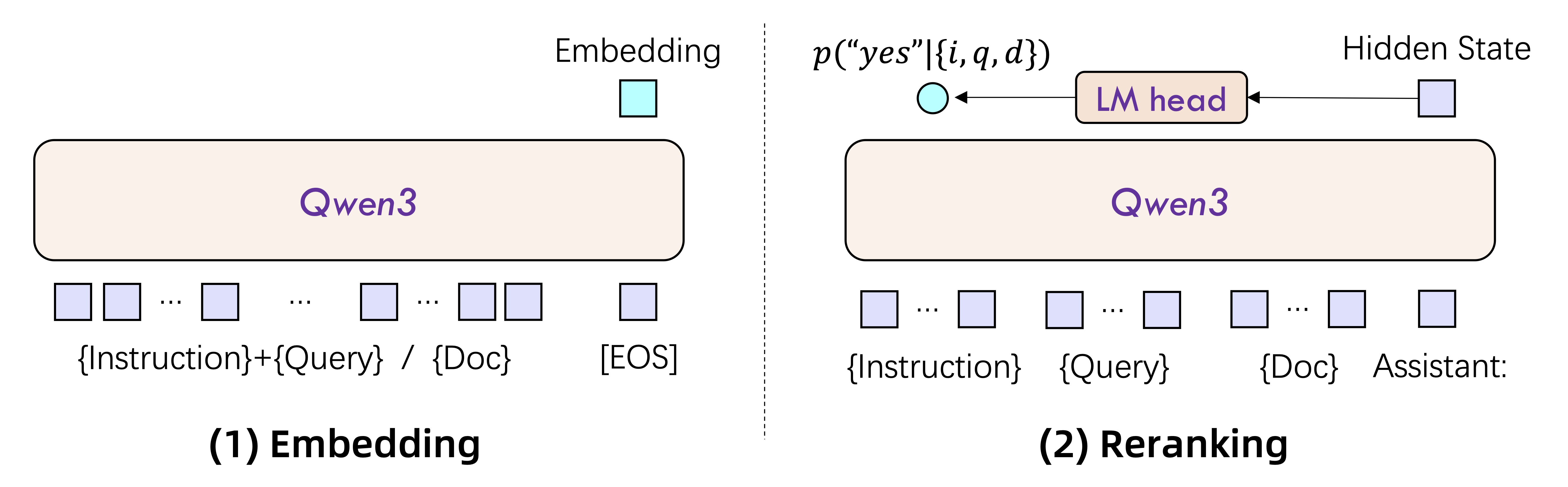}
\caption{Model architecture of Qwen3-Embedding (left) and Qwen3-Reranker (right).}
\label{fig:enter-label}
\end{figure}

\section{Model Architecture}
\label{sec:model_arc}

The core idea behind embedding and reranking models is to evaluate relevance in a task-aware manner. Given a query $q$ and a document $d$, embedding and reranking models assess their relevance based on a similarity criterion defined by instruction $I$. To enable the models for task-aware relevance estimation, training data is often organized as $\{I_i, q_i, d_i^+, d_{i,1}^-, \cdots, d_{i, n}^-\}$, where $d_i^+$ represents a positive (relevant) document for query $q_i$, and $d_{i,j}^-$ are negative (irrelevant) documents. Training the model on diverse text pairs broadens its applicability to a range of downstream tasks, including retrieval, semantic textual similarity, classification, and clustering.

\paragraph{Architecture} The Qwen3 embedding and reranking models are built on the dense version of Qwen3 foundation models and are available in three sizes: 0.6B, 4B, and 8B parameters. We initialize these models using the Qwen3 foundation models to leverage their capabilities in text modeling and instruction following. The model layers, hidden size, and context length for each model configuration are detailed in Table \ref{tab:table-model-arc}.

\paragraph{Embedding Models} For text embeddings, we utilize LLMs with causal attention, appending an \verb|[EOS]| token at the end of the input sequence. The final embedding is derived from the hidden state of the last layer corresponding to this \verb|[EOS]| token.

To ensure embeddings follow instructions during downstream tasks, we concatenate the instruction and the query into a single input context, while leaving the document unchanged before processing with LLMs. The input format for queries is as follows:
\begin{promptblock}
\begin{verbatim}
{Instruction} {Query}<|endoftext|>
\end{verbatim}
\end{promptblock}

\paragraph{Reranking Models} To more accurately evaluate text similarity, we employ LLMs for point-wise reranking within a single context. Similar to the embedding model, to enable instruction-following capability, we include the instruction in the input context. We use the LLM chat template and frame the similarity assessment task as a binary classification problem. The input to LLMs adheres to the template shown below:
\begin{promptblock}
\begin{verbatim}
<|im_start|>system
Judge whether the Document meets the requirements based on the Query and the Instruct provided. Note that the answer can only be "yes" or "no".<|im_end|>
<|im_start|>user
<Instruct>: {Instruction}
<Query>: {Query}
<Document>: {Document}<|im_end|>
<|im_start|>assistant
<think>\n\n</think>\n\n
\end{verbatim}
\end{promptblock}

To calculate the relevance score based on the given input, we assess the likelihood of the next token being "yes" or "no." This is expressed mathematically as follows:
\begin{equation*}
    \textrm{score}(q,d) = \frac{e^{P(\text{yes}|I,q,d)}}{e^{P(\text{yes}|I,q,d)}+e^{P(\text{no}|I,q,d)}}
\end{equation*}

\input{tables/model_arc}

%% file: tables/model_arc.tex
\begin{table}[t]
\centering
\resizebox{0.95\textwidth}{!}{%
\begin{tabular}{c|l|r|ccccc}
\toprule
Model Type & Models & Size & Layers & \begin{tabular}[c]{@{}c@{}}Sequence \\ Length\end{tabular} & \begin{tabular}[c]{@{}c@{}}Embedding \\ Dimension\end{tabular} & \begin{tabular}[c]{@{}c@{}}MRL \\ Support\end{tabular} & \begin{tabular}[c]{@{}c@{}}Instruction \\ Aware\end{tabular} \\
\midrule
\multirow{4}{*}{Text Embedding} & Qwen3-Embedding-0.6B & 0.6B & 28 & 32K & 1024 & Yes & Yes \\
 & Qwen3-Embedding-4B & 4B & 36 & 32K & 2560 & Yes & Yes \\
 & Qwen3-Embedding-8B & 8B & 36 & 32K & 4096 & Yes & Yes \\
 \midrule
\multirow{4}{*}{Text Reranking} & Qwen3-Reranker-0.6B & 0.6B & 28 & 32K & - & - & Yes \\
 & Qwen3-Reranker-4B & 4B & 36 & 32K & - & - & Yes \\
 & Qwen3-Reranker-8B & 8B & 36 & 32K & - & - & Yes \\
 \bottomrule
\end{tabular}%
}
\caption{Model architecture of Qwen3 Embedding models. ``MRL Support'' indicates whether the embedding model supports custom dimensions for the final embedding. ``Instruction Aware'' notes whether the embedding or reranker model supports customizing the input instruction according to different tasks.}
\label{tab:table-model-arc}
\end{table}

%% file: contents/train.tex
\begin{figure}
\centering
\includegraphics[width=\linewidth]{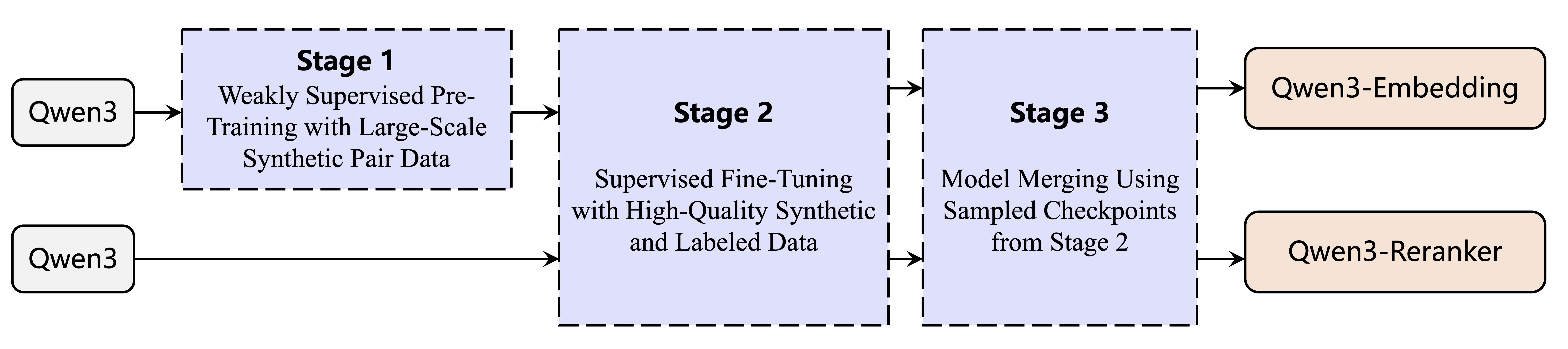}
\caption{Training pipeline of Qwen3 Embedding and Reranking models.}
\label{fig:pipeline}
\end{figure}

\section{Models Training}

In this section, we describe the multi-stage training pipeline adopted and present the key elements of this training recipe, including training objective, training data synthesis, and filtering of high-quality training data.

\subsection{Training Objective}
Before introducing our training pipeline, we first outline the optimized loss functions used for the embedding and reranking models during the training process. For the embedding model, we utilize an improved contrastive loss based on the InfoNCE framework~\citep{oord2018representation}. Given a batch of $N$ training instances, the loss is defined as:
\begin{equation}
\label{loss:embedding}
    L_\textrm{embedding} = - \frac{1}{N} \sum_i^N \log\frac{e^{(s(q_i, d_i^+)/\tau)}}{Z_i},
\end{equation}
where $s(\cdot, \cdot)$ is a similarity function (we use cosine similarity), $\tau$ is a temperature parameter, and $Z_i$ is the normalization factor that aggregates the similarity scores of the positive pair against various negative pairs:
\begin{equation*}
    Z_i = e^{(s(q_i, d_i^+) / \tau)}  + \sum_k^K m_{ik}e^{(s(q_i, d_{i,k}^-)/\tau)} + \sum_{j\neq i} m_{ij}e^{(s(q_i, q_j) / \tau)} + \sum_{j\neq i} m_{ij}e^{(s(d_i^+, d_j) / \tau)} + \sum_{j\neq i} m_{ij}e^{(s(q_i, d_j) / \tau)} 
\end{equation*}
where these terms represent similarities with: (1) the positive document $d_i^+$, (2) $K$ hard negatives $d_{i,k}^-$, (3) other in-batch queries $q_j$, (4) other in-batch documents $d_j$ compared against the positive document $d_i^+$. (5) other in-batch documents $d_j$ compared against the query $q_i$. The mask factor $m_{ij}$ is designed to mitigate the impact of false negatives and is defined as:
\begin{equation*}
 m_{ij} = 
\begin{cases} 
0 & \text{if } s_{ij} > s(q_i, d_i^+) + 0.1 \text{ or } d_j == d_i^+, \\
1 & \text{otherwise},
\end{cases}
\end{equation*}
among which $ s_{ij} $ is the corresponding score of $ q_i, d_j $ or $ q_i, q_j $. 

For the reranking model, we optimize the Supervised Fine-Tuning (SFT) loss defined as:
\begin{equation}
\label{loss:reranking}
        L_\textrm{reranking} = -\log p(l|\mathcal{P}(q,d)),
\end{equation}
where $p(\cdot|*)$ denotes the probability assigned by LLM. The label $l$ is ``yes'' for positive documents and ``no'' for negatives. This loss function encourages the model to assign higher probabilities to correct labels, thereby improving the ranking performance.

\subsection{Multi-stage Training}
The multi-stage training approach is a common practice for training text embedding models~\citep{li2023generaltextembeddingsmultistage,wang2024textembeddingsweaklysupervisedcontrastive,chen-etal-2024-m3}. This strategy typically begins with initial training on large-scale, semi-supervised data that includes noise, followed by fine-tuning using smaller, high-quality supervised datasets. This two-step process enhances the performance and generalization capabilities of embedding models. Large-scale weakly supervised training data contribute significantly to the model's generalization, while fine-tuning with high-quality data in subsequent stages further improves model performance. Both stages of training for embedding models utilize the optimization objective defined in Equation \ref{loss:embedding}, whereas the reranking model training employs the loss function defined in Equation \ref{loss:reranking} as the optimization target.

Building upon the existing multi-stage training framework, the Qwen3 Embedding series introduces the following key innovations:
\begin{itemize}
    \item Large-Scale Synthetic Data-Driven Weak Supervision Training: Unlike previous works (e.g., GTE, E5, BGE models), where weakly supervised training data are primarily collected from open-source communities such as Q$\&$A forums or academic papers, we propose leveraging the text understanding and generation capabilities of foundation models to synthesize pair data directly. This approach allows for arbitrary definition of various dimensions of the desired pair data, such as task, language, length, and difficulty within the synthesis prompts. Compared to data collection from open-domain sources, foundation model-driven data synthesis offers greater controllability, enabling precise management of the quality and diversity of the generated data, particularly in low-resource scenarios and languages.
    \item High-Quality Synthetic Data Utilization in Supervised Fine Tuning: Due to the exceptional performance of the Qwen3 Foundation model, the synthesized data is of notably high quality. Therefore, in the second stage of supervised training, selective incorporation of this high-quality synthetic data further enhances the overall model performance and generalization capabilities.
    \item Model Merging: Inspired by previous work~\citep{li2024improving}, after completing the supervised fine-tuning, we applied a model merging technique based on spherical linear interpolation (slerp). This technique involves merging multiple model checkpoints saved during the fine-tuning process. This step aims to boost the model’s robustness and generalization performance across various data distributions.
\end{itemize}
It is important to note that the reranking model's training process does not include a first-stage weakly supervised training phase.

\subsection{Synthetic Dataset}
To create a robust synthetic dataset for training models on various similarity tasks, we generate diverse text pairs spanning categories such as retrieval, bitext mining, classification, and semantic textual similarity (STS). The quality of these synthetic data pairs is ensured by utilizing the Qwen3-32B model as the foundational model for data synthesis. We have designed a diverse prompting strategy to improve the variety and authenticity of the generated data. For instance, in the text retrieval task, we synthesize data using the multilingual pre-training corpus from Qwen3. During the data synthesis process, specific roles are assigned to each document to simulate potential users querying that document. This injection of user perspectives enhances the diversity and realism of the synthetic queries. Specifically, we utilize a retrieval model to identify the top five role candidates for each document from a role library and present these documents along with their role candidates to the prompt. This guides the model in outputting the most suitable role configuration for query generation. Moreover, the prompt incorporates various dimensions such as query type (e.g., keyword, factual, summary, judgment), query length, difficulty, and language. This multidimensional approach ensures the quality and diversity of the synthetic data.

Finally, we create a total of approximately 150 million pairs of multi-task weak supervision training data. Our experiments reveal that the embedding model trained with these synthetic data performs exceptionally well in downstream evaluations, particularly surpassing many previously supervised models in the MTEB Multilingual benchmarks. This motivates us to filter the synthetic data to identify high-quality pairs for inclusion in a second stage of supervised training. We employ a simple cosine similarity calculation to select data pairs, retaining those with a cosine similarity greater than 0.7 from randomly sampled data. Ultimately, approximately 12 million high-quality supervised training data pairs are selected for further training.

%% file: contents/evaluation.tex
\input{tables/mteb_multilingual}

\section{Evaluation}
We conduct comprehensive and fair evaluations across multiple benchmarks to assess the capabilities of Qwen3 Embedding models.

\subsection{Settings}
\label{sec:evaluation_settings}
For the text embedding models, we utilize the Massive Multilingual Text Embedding Benchmark (MMTEB)~\citep{enevoldsen2025mmteb} for evaluation. MMTEB is a large-scale, community-driven expansion of MTEB~\citep{muennighoff-etal-2023-mteb}, covering over 500 quality-controlled evaluation tasks across more than 250 languages. In addition to classic text tasks such as as a variety of retrieval, classification, and semantic textual similarity, MMTEB includes a diverse set of challenging and novel tasks, such as instruction following, long-document retrieval, and code retrieval, representing the largest multilingual collection of evaluation tasks for embedding models to date. Our MMTEB evaluations encompass 216 individual evaluation tasks, consisting of 131 tasks for MTEB (Multilingual) \citep{enevoldsen2025mmteb}, 41 tasks for MTEB (English, v2) \citep{muennighoff-etal-2023-mteb}, 32 tasks for CMTEB \citep{xiao2024cpack}, and 12 code retrieval tasks for MTEB (Code) \citep{enevoldsen2025mmteb}.

Moreover, we select a series of text retrieval tasks to assess the text reranking capabilities of our models. We explore three types of retrieval tasks: (1) Basic Relevance Retrieval, categorized into English, Chinese, and Multilingual, evaluated on MTEB \citep{muennighoff-etal-2023-mteb}, CMTEB \citep{xiao2024cpack}, MMTEB \citep{enevoldsen2025mmteb}, and MLDR \citep{chen-etal-2024-m3}, respectively; (2) Code Retrieval, evaluated on MTEB-Code \citep{enevoldsen2025mmteb}, which comprises only code-related retrieval data.; and (3) Complex Instruction Retrieval, evaluated on FollowIR \citep{weller2024followir}.

\paragraph{Compared Methods}
We compare our models with the most prominent open-source text embedding models and commercial API services. The open-source models include the GTE \citep{li2023generaltextembeddingsmultistage,zhang-etal-2024-mgte}, E5 \citep{wang2024textembeddingsweaklysupervisedcontrastive}, and BGE \citep{xiao2024cpack} series, as well as NV-Embed-v2 \citep{lee2025nvembed}, GritLM-7B \cite{muennighoff2025generative}. The commercial APIs evaluated are text-embedding-3-large from OpenAI, Gemini-embedding from Google, and Cohere-embed-multilingual-v3.0.
For reranking, we compare with the rerankers of jina\footnote{
\url{https://hf.co/jinaai/jina-reranker-v2-base-multilingual}
}, mGTE \citep{zhang-etal-2024-mgte} and BGE-m3 \citep{chen-etal-2024-m3}.

\input{tables/mteb_ezc}

\subsection{Main Results}

\paragraph{Embedding} In Table \ref{tab:model_comparison}, we present the evaluation results on MMTEB~\citep{enevoldsen2025mmteb}, which comprehensively covers a wide range of embedding tasks across multiple languages.
Our Qwen3-Embedding-4B/8B models achieve the best performance, and our smallest model, Qwen3-Embedding-0.6B, only lags behind the best-performing baseline method (Gemini-Embedding), despite having only 0.6B parameters.
In Table \ref{tab:mteb-ezc}, we present the evaluation results on MTEB (English, v2) \citep{muennighoff-etal-2023-mteb}, CMTEB~\citep{xiao2024cpack}, and MTEB (Code)~\citep{enevoldsen2025mmteb}.
The scores reflect similar trends as MMTEB, with our Qwen3-Embedding-4B/8B models consistently outperforming others. Notably, the Qwen3-Embedding-0.6B model ranks just behind the Gemini-Embedding, while being competitive with the gte-Qwen2-7B-instruct.

\paragraph{Reranking} In Table \ref{tab:main-rerank}, we present the evaluation results on various reranking tasks (\S\ref{sec:evaluation_settings}). We utilize the Qwen3-Embedding-0.6B model to retrieve the top-100 candidates and then apply different reranking models for further refinement. This approach ensures a fair evaluation of the reranking models. Our results indicate that all three Qwen3-Reranker models enhance performance compared to the embedding model and surpass all baseline reranking methods, with Qwen3-Reranker-8B achieving the highest performance across most tasks.

\input{tables/rerank_all}

\subsection{Analysis}

To further analyze and explore the key elements of the Qwen3 Embedding model training framework, we conduct an analysis from the following dimensions:

\noindent \textbf{Effectiveness of Large-Scale Weakly Supervised Pre-Training}
We first analyze the effectiveness of the large-scale weak supervised training stage for the embedding models. As shown in Table \ref{tab:mteb-analysis}, the Qwen3-Embedding-0.6B model trained solely on synthetic data (without subsequent training stages, as indicated in the first row) achieves reasonable and strong performance compared to the final Qwen3-Embedding-0.6B model (as shown in the last row). If we further remove the weak supervised training stage (i.e., without synthetic data training, as seen in the second row), the final performance shows a clear decline. This indicates that the large-scale weak supervised training stage is crucial for achieving superior performance.

\noindent \textbf{Effectiveness of Model Merging}
Next, we compare the performance differences arising from the model merging stage. As shown in Table \ref{tab:mteb-analysis}, the model trained without model merging techniques (the third row, which uses data sampling to balance various tasks) performs considerably worse than the final Qwen3-Embedding-0.6B model (which employs model merging, as shown in the last row). This indicates that the model merging stage is also critical for developing strong models.

\input{tables/mteb_analysis.tex}

%% file: tables/mteb_multilingual.tex
\begin{table}[h]
\centering \setlength{\tabcolsep}{3pt}
\resizebox{\textwidth}{!}{
\begin{tabular}{lr|c|c|ccccccccc}
\toprule
\textbf{Model}                 & \textbf{Size} & 
\begin{tabular}[c]{@{}c@{}} \textbf{Mean} \\ \textbf{(Task)} \end{tabular} & 
\begin{tabular}[c]{@{}c@{}} \textbf{Mean} \\ \textbf{(Type)} \end{tabular} &
\begin{tabular}[c]{@{}c@{}}  Bitext \\ Mining
\end{tabular} & 
\begin{tabular}[c]{@{}c@{}}  Class- \\ ification
\end{tabular} &
\begin{tabular}[c]{@{}c@{}}  Clus- \\ tering
\end{tabular} &
\begin{tabular}[c]{@{}c@{}}  Inst. \\ Retrieval
\end{tabular} &
\begin{tabular}[c]{@{}c@{}}  Multilabel \\ Class.
\end{tabular} &
\begin{tabular}[c]{@{}c@{}}  Pair \\ Class.
\end{tabular} &
Rerank & Retrieval & STS \\
\midrule
\multicolumn{13}{c}{\textbf{Selected Open-Source Models}} \\
\midrule
NV-Embed-v2 & 7B & 56.29 & 49.58 & 57.84 & 57.29 & 40.80 & 1.04 & 18.63 & 78.94 & 63.82 & 56.72 & 71.10 \\ 
GritLM-7B & 7B & 60.92 & 53.74 & 70.53 & 61.83 & 49.75 & 3.45 & 22.77 & 79.94 & 63.78 & 58.31 & 73.33 \\
BGE-M3                & 0.6B & 59.56 & 52.18 & 79.11 & 60.35 & 40.88 & -3.11 & 20.1 & 80.76 & 62.79 & 54.60 & 74.12 \\
multilingual-e5-large-instruct & 0.6B & 63.22 & 55.08 & 80.13 & 64.94 & 50.75 & -0.40 & 22.91 & 80.86 & 62.61 & 57.12 & 76.81 \\
gte-Qwen2-1.5B-instruct & 1.5B & 59.45 & 52.69 & 62.51 & 58.32 & 52.05 & 0.74 & 24.02 & 81.58 & 62.58 & 60.78 & 71.61 \\
gte-Qwen2-7b-Instruct & 7B   & 62.51 & 55.93 & 73.92 & 61.55 & 52.77 & 4.94 & 25.48 & 85.13 & 65.55 & 60.08 & 73.98 \\
\midrule
\multicolumn{13}{c}{\textbf{Commercial APIs}} \\
\midrule
text-embedding-3-large &  -    & 58.93 & 51.41 & 62.17 & 60.27 & 46.89 & -2.68 & 22.03 & 79.17 & 63.89 & 59.27 & 71.68 \\
Cohere-embed-multilingual-v3.0 & - &  61.12 & 53.23 & 70.50 & 62.95 & 46.89 & -1.89 & 22.74 & 79.88 & 64.07 & 59.16 & 74.80 \\
Gemini Embedding      & -    & 68.37 & 59.59 & 79.28 & 71.82 & 54.59 & 5.18 & \bf{29.16} & 83.63 & 65.58 & 67.71 & 79.40 \\
\midrule
\multicolumn{13}{c}{\textbf{Qwen3 Embedding Models}} \\
\midrule
\bf Qwen3-Embedding-0.6B     & 0.6B    & 64.33 & 56.00 & 72.22 & 66.83 & 52.33 & 5.09 & 24.59 & 80.83 & 61.41 & 64.64 & 76.17 \\
\bf Qwen3-Embedding-4B     & 4B    & 69.45 & 60.86 & 79.36 & 72.33 & 57.15 & \bf{11.56} & 26.77 & 85.05 & 65.08 & 69.60 & 80.86 \\
\bf Qwen3-Embedding-8B     & 8B    & \bf{70.58} & \bf{61.69} & \bf{80.89} & \bf{74.00} & \bf{57.65} & 10.06 &  28.66 & \bf{86.40} & \bf{65.63} & \bf{70.88} & \bf{81.08} \\
\bottomrule
\end{tabular}
}
\caption{Performance on MTEB Multilingual \citep{enevoldsen2025mmteb}. For compared models, the scores are retrieved from MTEB online \href{https://huggingface.co/spaces/mteb/leaderboard}{leaderboard} on June 4th, 2025.}
\label{tab:model_comparison}
\end{table}

%% file: tables/mteb_ezc.tex
\begin{table}
\setlength{\tabcolsep}{3pt}
\resizebox{\textwidth}{!}{
\begin{tabular}{lr|c|cc|cc|c}
\toprule
Model                       & Size & Dim & \multicolumn{2}{c}{MTEB (Eng, v2)} & \multicolumn{2}{c}{CMTEB} & MTEB (Code) \\ 
\midrule
                           &      &     & Mean (Task) & Mean (Type) & Mean (Task) & Mean (Type) &              \\ 
\midrule
\multicolumn{8}{c}{\textbf{Selected Open-Source Models}} \\
\midrule
NV-Embed-v2               & 7B   & 4096  & 69.81   & 65.00        & 63.0        & 62.0        & -         \\ 
GritLM-7B    & 7B &  4096 & 67.07  &  63.22  & - & - & 73.6$^\alpha$ \\
multilingual-e5-large-instruct & 0.6B & 1024 & 65.53    & 61.21        & -       & -        & 65.0$^\alpha$         \\ 
gte-Qwen2-1.5b-instruct   & 1.5B & 1536  &    67.20 & 63.26    & 67.12 & 67.79        & -        \\ 
gte-Qwen2-7b-instruct     & 7B   & 3584  & 70.72        & 65.77        & 71.62    & 72.19        & 56.41$^\gamma$ \\ 
\midrule
\multicolumn{8}{c}{\textbf{Commercial APIs}} \\
\midrule
text-embedding-3-large    & -    & 3072    & 66.43 & 62.15        & -        & -        & 58.95$^\gamma$         \\
cohere-embed-multilingual-v3.0   & -    & 1024    & 66.01   & 61.43  & -        & -        & 51.94$^\gamma$         \\ 
Gemini Embedding           & -    & 3072    & 73.30        & 67.67        & -        & -        & 74.66$^\gamma$         \\ 
\midrule
\multicolumn{8}{c}{\textbf{Qwen3 Embedding Models}} \\
\midrule
\bf Qwen3-Embedding-0.6B      & 0.6B & 1024 & 70.70        & 64.88        & 66.33        & 67.44        & 75.41         \\ 
\bf Qwen3-Embedding-4B        & 4B   & 2560 & 74.60 & 68.09 &   72.26 & 73.50 & 80.06  \\ 
\bf Qwen3-Embedding-8B        & 8B   & 4096 & \bf{75.22} & \bf{68.70} & \bf{73.83} & \bf{75.00} & \bf{80.68}        \\ 
\bottomrule
\end{tabular}
}
\caption{Performance on MTEB Engilish, MTEB Chinese, MTEB Code. $^\alpha$Taken from \citep{enevoldsen2025mmteb}. $^\gamma$Taken from \citep{lee2025gemini}. For other compared models, the scores are retrieved from MTEB online \href{https://huggingface.co/spaces/mteb/leaderboard}{leaderboard} on June 4th, 2025.}
\label{tab:mteb-ezc}
\end{table}

%% file: tables/rerank_all.tex
\begin{table}
\centering
\resizebox{\linewidth}{!}{
\setlength{\tabcolsep}{3pt}
\begin{tabular}{lc|cccccc}
\toprule
&& \multicolumn{4}{c}{Basic Relevance Retrieval} &  &  \\
\cmidrule(lr){3-6}\cmidrule(lr){7-7}\cmidrule(lr){8-8}
Model & Param & MTEB-R & CMTEB-R & MMTEB-R & MLDR & MTEB-Code & FollowIR  \\
\midrule
\bf Qwen3-Embedding-0.6B & 0.6B &  61.82 & 71.02 & 64.64 & 50.26 & 75.41 & 5.09 \\
\midrule
Jina-multilingual-reranker-v2-base & 0.3B & 58.22 & 63.37 & 63.73 & 39.66 & 58.98 & -0.68  \\
gte-multilingual-reranker-base & 0.3B & 59.51 & 74.08 & 59.44 & 66.33 & 54.18 & -1.64  \\
BGE-reranker-v2-m3 & 0.6B & 57.03 & 72.16 & 58.36 & 59.51 & 41.38 & -0.01 \\
\midrule
\bf Qwen3-Reranker-0.6B & 0.6B  & 65.80 & 71.31 & 66.36 & 67.28 & 73.42 & 5.41 \\
\bf Qwen3-Reranker-4B & 4B  & \bf{69.76} & 75.94 & 72.74 & 69.97 & 81.20 & \bf{14.84} \\ 
\bf Qwen3-Reranker-8B & 8B  & 69.02 & \bf{77.45} & \bf{72.94} & \bf{70.19} & \bf{81.22} & 8.05 \\ 
\bottomrule
\end{tabular}}
\caption{
Evaluation results for reranking models. We use the retrieval subsets of MTEB(eng, v2), MTEB(cmn, v1) and MMTEB, which are MTEB-R, CMTEB-R and MMTEM-R.
The rest are all retrieval tasks.
All scores are our runs based on the retrieval top-$100$ results from the first row.
}
\label{tab:main-rerank}
\end{table}

%% file: tables/mteb_analysis.tex
\begin{table}
\setlength{\tabcolsep}{3pt}
\resizebox{\textwidth}{!}{
\begin{tabular}{l|c|c|c|c}
\toprule
Model & MMTEB & MTEB (Eng, v2) & CMTEB & MTEB (Code, v1) \\ 
\midrule
\bf Qwen3-Embedding-0.6B w/ only synthetic data & 58.49 & 60.63 & 59.78 & 66.79 \\
\bf Qwen3-Embedding-0.6B w/o synthetic data & 61.21 & 65.59 & 63.37 & 74.58 \\
\bf Qwen3-Embedding-0.6B w/o model merge &  62.56 & 68.18 & 64.76 & 74.89 \\
\midrule
\bf Qwen3-Embedding-0.6B & 64.33 & 70.70 & 66.33 & 75.41 \\
\bottomrule
\end{tabular}
}
\caption{Performance (mean task) on MMTEB, MTEB(eng, v2), CMTEB and MTEB(code, v1) for Qwen3-Embedding-0.6B model with different training setting.}
\label{tab:mteb-analysis}
\end{table}

%% file: tables/train_data.tex
\begin{table*}[h!]
\centering
\resizebox{\textwidth}{!}{
\begin{tabular}{l|c|c}
\toprule
\textbf{Stage} & \textbf{Dataset} & \textbf{Size} \\
\midrule
Weakly Supervised Pre-Training & Synthetic Data & 
    \begin{tabular}{@{}c@{}}
    $\sim 150\text{M}$ \\ 
    \end{tabular} \\
\midrule
Supervised Fine Tuning & 
    \begin{tabular}{@{}c@{}}
    MS MARCO, NQ, HotpotQA, NLI, \\
    Dureader, T$^2$-Ranking, SimCLUE, \\
    MIRACL, MLDR, Mr.TyDi, \\
    Multi-CPR, CodeSearchNet .etc \\ 
    + High-quality Synthetic Data
    \end{tabular} &
    \begin{tabular}{@{}c@{}}
    Labeled Data: $\sim 7\text{M}$ \\ 
    Synthetic Data: $\sim 12\text{M}$
    \end{tabular} \\
\bottomrule
\end{tabular}}
\caption{Statistics of training data utilized at each stage.}
\label{tab:ft-data}
\end{table*}

%% file: tables/mteb_en.tex
\begin{table*}[ht]
\centering
\resizebox{\textwidth}{!}{
\begin{tabular}{lccc|cccccccc}
\toprule

\bf MTEB(eng, v2) & \bf Param & 
\begin{tabular}[c]{@{}c@{}} \textbf{Mean} \\ \textbf{(Task)}
\end{tabular} &
\begin{tabular}[c]{@{}c@{}} \textbf{Mean} \\ \textbf{(Type)}
\end{tabular} &
\begin{tabular}[c]{@{}c@{}}  Class- \\ ification
\end{tabular} &
\begin{tabular}[c]{@{}c@{}}  Clus- \\ tering
\end{tabular} &
\begin{tabular}[c]{@{}c@{}}  Pair \\ Class.
\end{tabular} &
Rerank & Retrieval & STS & Summ. \\
\midrule
multilingual-e5-large-instruct & 0.6B & 65.53 &	61.21 & 75.54 & 49.89 & 86.24 & 48.74 & 53.47 & 84.72 & 29.89 \\
NV-Embed-v2 & 7.8B  & 69.81 & 65.00 & 87.19 & 47.66 & 88.69 & 49.61 & 62.84 & 83.82 & 35.21 \\
GritLM-7B & 7.2B  & 67.07 & 63.22	& 81.25	 & 50.82 & 87.29 & 49.59 & 54.95 & 83.03 & 35.65  \\
gte-Qwen2-1.5B-instruct  & 1.5B  & 67.20 & 63.26 & 85.84 & 53.54 & 87.52 & 49.25 & 50.25 & 82.51 & 33.94 \\
stella\_en\_1.5B\_v5 & 1.5B & 69.43 & 65.32 &  89.38	& 57.06	& 88.02	& 50.19	& 52.42	& 83.27	& 36.91 \\
gte-Qwen2-7B-instruct  & 7.6B  &  70.72 & 65.77 & 88.52 & 58.97 & 85.9 & 50.47 & 58.09 & 82.69 & 35.74 \\
gemini-embedding-exp-03-07 & - & 73.3 & 67.67 & 90.05 & 59.39 & 87.7 & 48.59 & 64.35 & 85.29 & 38.28 \\
\midrule
Qwen3-Embedding-0.6B & 0.6B  & 70.70 & 64.88 & 85.76 & 54.05 & 84.37 & 48.18 & 61.83 & 86.57 & 33.43\\
Qwen3-Embedding-4B & 4B  & 74.60 & 68.09 & 89.84 & 57.51 & 87.01 & 50.76 & 68.46 & 88.72 & 34.39\\
Qwen3-Embedding-8B & 8B  & 75.22 & 68.70 & 90.43 & 58.57 & 87.52 & 51.56 & 69.44 & 88.58 & 34.83 \\
\bottomrule
\end{tabular}}
\caption{
Results on MTEB(eng, v2) \citep{muennighoff-etal-2023-mteb}. We compare models from the online leaderboard.
}
\label{tab:mteb-en}
\end{table*}

%% file: tables/mteb_zh.tex
\begin{table*}[h]
\centering
\resizebox{\textwidth}{!}{
\begin{tabular}{lccc|cccccc}
\toprule[1pt]

\bf MTEB(cmn, v1) & \bf Param & 
\begin{tabular}[c]{@{}c@{}} \textbf{Mean} \\ \textbf{(Task)}
\end{tabular} &
\begin{tabular}[c]{@{}c@{}} \textbf{Mean} \\ \textbf{(Type)}
\end{tabular} &
\begin{tabular}[c]{@{}c@{}}  Class- \\ ification
\end{tabular} &
\begin{tabular}[c]{@{}c@{}}  Clus- \\ tering
\end{tabular} &
\begin{tabular}[c]{@{}c@{}}  Pair \\ Class.
\end{tabular} &
Rerank & Retrieval & STS \\
\midrule
multilingual-e5-large-instruct & 0.6B  & 58.08	& 58.24 & 69.80 & 48.23	& 64.52	& 57.45	 & 63.65 & 45.81 \\ 
gte-Qwen2-7B-instruct     & 7.6B  & 71.62 & 72.19 & 75.77 & 66.06 & 81.16 & 69.24 & 75.70 & 65.20  \\
gte-Qwen2-1.5B-instruct  &  1.5B  & 67.12 & 67.79	& 72.53 & 54.61 & 79.5 & 68.21 & 71.86 & 60.05 \\
\midrule
Qwen3-Embedding-0.6B & 0.6B  & 66.33 & 67.44 & 71.40 & 68.74 & 76.42 & 62.58 & 71.03 & 54.52 \\
Qwen3-Embedding-4B &  4B  & 72.26 & 73.50 & 75.46 & 77.89 & 83.34 & 66.05 & 77.03 & 61.26 \\
Qwen3-Embedding-8B &  8B  & 73.84 & 75.00 & 76.97 & 80.08 & 84.23 & 66.99 & 78.21 & 63.53 \\
\bottomrule[1pt]
\end{tabular}}
\caption{
Results on C-MTEB \citep{xiao2024cpack} (MTEB(cmn, v1).
}
\label{tab:mteb-zh}
\end{table*}

%% file: tables/mteb_code.tex
\begin{table*}[h]
\centering
\setlength{\tabcolsep}{2pt}
\resizebox{\textwidth}{!}{
\begin{tabular}{lc|ccccccccccccc}
\toprule
MTEB(Code, v1) &
Avg. &
\begin{tabular}{c} Apps \end{tabular} &
\begin{tabular}{c} COIR-\\CodeSearch-\\Net \end{tabular} &
\begin{tabular}{c} Code-\\Edit-\\Search\end{tabular} &
\begin{tabular}{c} Code-\\Feedback-\\MT \end{tabular} & 
\begin{tabular}{c} Code-\\Feedback-\\ST \end{tabular} & 
\begin{tabular}{c} Code-\\SearchNet-\\CCR \end{tabular} &
\begin{tabular}{c} Code-\\SearchNet \end{tabular} &
\begin{tabular}{c}  Code-\\Trans-\\Ocean-\\Contest \end{tabular} &
\begin{tabular}{c} Code-\\Trans-\\Ocean-DL \end{tabular} &
CosQA & 
\begin{tabular}{c}  Stack-\\Overflow-\\QA \end{tabular} &
\begin{tabular}{c} Synthetic-\\Text2SQL  \end{tabular} &

\\ \midrule
BGE$_\text{multilingual}$  & 62.04 & 22.93 & 68.14 & 60.48 & 60.52 & 76.70 & 73.23 & 83.43 & 86.84 & 32.64 & 27.93 & 92.93 & 58.67 \\
NV-Embed-v2 & 63.74 & 29.72 & 61.85 &73.96 & 60.27 & 81.72 & 68.82 & 86.61 & 89.14 & 33.40 & 34.82 & 92.36 & 60.90 \\
gte-Qwen2-7B-instruct & 62.17 & 28.39 & 71.79 & 67.06 & 57.66 & 85.15 & 66.24 & 86.96 & 81.83 & 32.17 & 31.26 & 84.34 & 53.22  \\
gte-Qwen2-1.5B-instruct & 61.98 & 28.91 & 71.56 & 59.60 & 49.92 & 81.92 & 72.08 & 91.08 & 79.02 & 32.73 & 32.23 & 90.27 & 54.49 \\
\midrule
BGE-M3 (Dense)   & 58.22 &  14.77 & 58.07 & 59.83 & 47.86 & 69.27 & 53.55 & 61.98 & 86.22 & 29.37 & 27.36 & 80.71 & 49.65 \\
Jina-v3 & 58.85 & 28.99 & 67.83 & 57.24 & 59.66 & 78.13 & 54.17 & 85.50 & 77.37 & 30.91 & 35.15 & 90.79 & 41.49 \\
\midrule
Qwen3-Embedding-0.6B & 75.41 & 75.34 & 84.69 & 64.42 & 90.82 & 86.39 & 91.72 & 91.01 & 86.05 & 31.36 & 36.48 & 89.99 & 76.74 \\
Qwen3-Embedding-4B &  80.06 & 89.18 & 87.93 & 76.49 & 93.21 & 89.51 & 95.59 & 92.34 & 90.99 & 35.04 & 37.98 & 94.32 & 78.21 \\
Qwen3-Embedding-8B & 80.68 & 91.07 & 89.51 & 76.97 & 93.70 & 89.93 & 96.35 & 92.66 & 93.73 & 32.81 & 38.04 & 94.75 & 78.75 \\
\midrule
Qwen3-Reranker-0.6B & 73.42 & 69.43 & 85.09 & 72.37 & 83.83 & 78.05 & 94.76 & 88.8 & 84.69 & 33.94 & 36.83 & 93.24 & 62.48 \\
Qwen3-Reranker-4B &  81.20 & 94.25 & 90.91 & 82.53 & 95.25 & 88.54 & 97.58 & 92.48 & 93.66 & 36.78 & 35.14 & 97.11 & 75.06 \\
Qwen3-Reranker-8B & 81.22 & 94.55 & 91.88 & 84.58 & 95.64 & 88.43 & 95.67 & 92.78 & 90.83 & 34.89 & 37.43 & 97.3 & 73.4 \\
\bottomrule
\end{tabular}
}
\caption{
Performance on \texttt{MTEB(Code, v1)}~\citep{enevoldsen2025mmteb}. We report nDCG@10 scores.
}
\label{tab:mteb-code_results}
\end{table*}